\documentclass{article}

\usepackage{microtype}
\usepackage{graphicx}
\usepackage{subfigure}
\usepackage{booktabs} 

\usepackage{hyperref}

\usepackage[accepted]{icml2025}

\usepackage{amsmath}
\usepackage{amssymb}
\usepackage{mathtools}
\usepackage{amsthm}
\usepackage{graphicx}
\usepackage{dblfloatfix}

\usepackage[capitalize,noabbrev]{cleveref}

\theoremstyle{plain}

\theoremstyle{definition}

\theoremstyle{remark}

\usepackage[textsize=tiny]{todonotes}

\icmltitlerunning{e-SimFT: Alignment of Generative Models with Simulation Feedback for Pareto-Front Design Exploration}

\begin{document}

\twocolumn[
\icmltitle{e-SimFT: Alignment of Generative Models with Simulation Feedback for Pareto-Front Design Exploration}



\icmlsetsymbol{equal}{*}

\begin{icmlauthorlist}
\icmlauthor{Hyunmin Cheong}{yyy}
\icmlauthor{Mohammadmehdi Ataei}{yyy}
\icmlauthor{Amir Hosein Khasahmadi}{yyy}
\icmlauthor{Pradeep Kumar Jayaraman}{yyy}
\end{icmlauthorlist}

\icmlaffiliation{yyy}{Autodesk Research, Toronto, Ontario, Canada}

\icmlcorrespondingauthor{Hyunmin Cheong}{hyunmin.cheong@autodesk.com}
\icmlkeywords{Preference Fine-tuning, Multi-objective Alignment, Simulation Feedback, Engineering Design}

\vskip 0.3in
]



\printAffiliationsAndNotice{} 

\begin{abstract}
Deep generative models have recently shown success in solving complex engineering design problems where models predict solutions that address the design requirements specified as input. However, there remains a challenge in aligning such models for effective design exploration. For many design problems, finding a solution that meets all the requirements is infeasible. In such a case, engineers prefer to obtain a set of Pareto optimal solutions with respect to those requirements, but uniform sampling of generative models may not yield a useful Pareto front. To address this gap, we introduce a new framework for Pareto-front design exploration with simulation fine-tuned generative models. First, the framework adopts preference alignment methods developed for Large Language Models (LLMs) and showcases the first application in fine-tuning a generative model for engineering design. The important distinction here is that we use a simulator instead of humans to provide accurate and scalable feedback. Next, we propose epsilon-sampling, inspired by the epsilon-constraint method used for Pareto-front generation with classical optimization algorithms, to construct a high-quality Pareto front with the fine-tuned models. Our framework, named e-SimFT, is shown to produce better-quality Pareto fronts than existing multi-objective alignment methods.
\end{abstract}

\section{Introduction}

Generative artificial intelligence (AI) has made remarkable implications in many domains, especially where creative automation is greatly desired. One notable area is engineering design, where generative AI has the potential to help engineers develop solutions to their problems at a much faster pace than with the traditional design process. Such progress could bring significant innovation to real-world problems and therefore amplify AI's positive impact on our society.

Several efforts have been made to apply deep generative models to solve engineering design problems, as reviewed in \cite{regenwetter2022deep}. However, much of the prior work is limited to solving a problem with a fixed set of design requirements and cannot consider of different requirements that the user may provide. In contrast, recent work such as \cite{etesam2024deep} has developed a generative model that takes a set of design requirements as input and outputs a design solution conditioned on those requirements.

There remains a challenge in making use of such generative models in practice. For many design problems, finding a solution that meets all the specified requirements is often impossible. Even a highly capable generative model is unlikely to produce a perfect solution, given the problem's inherent complexity. In such a scenario, engineers could focus on finding solutions for a relatively more important subset of the requirements. Or preferably, they would like to obtain a set of Pareto optimal solutions (i.e., a Pareto front) with respect to the important requirements so that they can understand the trade-offs and compare alternative solutions.

This challenge highlights new research opportunities in two aspects. First, we need to align a generative model with respect to specific design requirements preferred by the engineer. Next, we need a method to effectively sample a generative model to produce a high-quality Pareto front. 

\begin{figure*}[h!]
    \centering
    \includegraphics[width=0.87\linewidth]{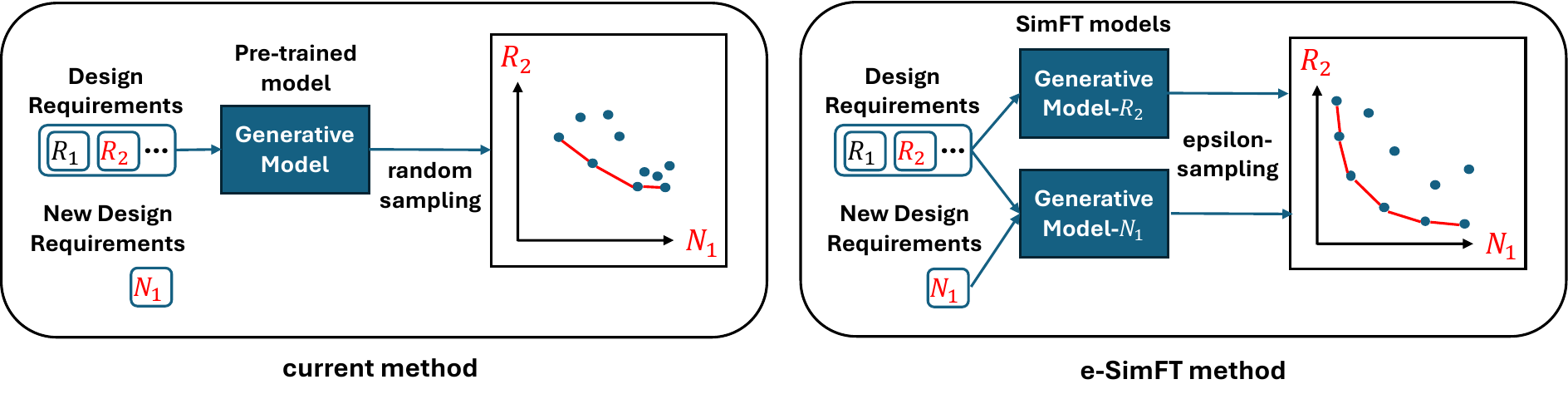}
    \caption{Randomly sampling a generative model (for engineering design) may not yield a good Pareto front with respect to the design requirements of interest. Also, an engineer could be interested in a new design requirement that the current model is not conditioned on to generate design solutions. We address these issues with SimFT methods -- using simulation feedback to fine-tune a generative model with respect to specific design requirements, including new ones not seen by the model, and proposing a new sampling strategy inspired by the epsilon-constraint method to create a high-quality Pareto front.}
    \label{fig:overview}
\end{figure*}

This work demonstrates that the alignment problem can be solved by adapting fine-tuning methods used for LLMs, but using \emph{simulation feedback}. Because a simulator can be used to evaluate a given solution with respect to design requirements of interest, we can use it to either generate fine-tuning data or compute rewards for Reinforcement Learning (RL). We show that different fine-tuning methods are more applicable and effective depending on the requirements. 

In addition, we propose \emph{epsilon-sampling}, which leverages the simulation fine-tuned models to produce a high-quality Pareto front. The technique is inspired by the epsilon-constraint method \cite{haimes1971bicriterion} used for constructing a Pareto front with gradient-based optimization. 

The main contributions of our work are as follows:

\begin{itemize}
    \item The first application of preference alignment methods to fine-tune a generative model in engineering design.
    
    \item \emph{SimFT}: Dataset generation and fine-tuning methods with simulator feedback to prioritize satisfying different design requirements with the solutions generated.
    
    \item \emph{e-SimFT}: A new sampling strategy called epsilon-sampling with simulation fine-tuned models to construct a high-quality Pareto-front.
\end{itemize}

The overview of our method is illustrated in \cref{fig:overview}. We showcase the performance improvements made with respect to specific design requirements using SimFT methods and various ablation studies to elucidate the nature of simulation fine-tuning. We also evaluate e-SimFT against several baselines including multi-objective alignment methods developed for LLMs and show its overall superior performance.  

\section{Related Work}

Preference alignment has gained notable attention, particularly in guiding LLMs to generate contents that align with the user's values or objectives. The primary approach to achieving this is through Reinforcement Learning with Human Feedback (RLHF) \cite{christiano2017deep}, and several successful applications have been reported for fine-tuning LLMs for different tasks \cite{ziegler2019fine, stiennon2020learning, ouyang2022training}, and extended to incorporate other feedback sources than humans \cite{leerlaif, liu2023rltf, jha2024rlsf, williams2024multi}. The most popular RL algorithm used for these methods is Proximal Policy Optimization (PPO) \cite{schulman2017proximal}. Another alternative is to directly fine-tune LLMs with a preference dataset without RL, e.g., \cite{hejna2023contrastive, azar2024general, ethayarajh2024kto}, and most notably Direct Preference Optimization (DPO) \cite{rafailov2024direct}. 

Because user preferences for LLMs are likely multidimensional with trade-offs, e.g., helpfulness vs. harmlessness, multi-objective alignment methods have been proposed to produce Pareto-front aligned models with respect to multiple preference criteria. Notable recent work includes Rewarded Soup \cite{rame2024rewarded}, multi-objective DPO \cite{zhou2024beyond}, Rewards-in-Context \cite{yang2024rewards}, controllable preference optimization \cite{guo2024controllable}, and Panacea \cite{zhong2024panacea}.

Note that the purpose of multi-objective alignment methods has a strong parallel with the purpose of multi-objective optimization methods \cite{deb2016multi} where the goal is to find models or solutions that constitute a high-quality Pareto front \cite{zitzler1998multiobjective}. Therefore, we were motivated to find inspirations from the techniques used in the latter domain such as the epsilon-constraint method \cite{haimes1971bicriterion} or non-dominated sorting \cite{deb2002fast}. 

\section{Problem}

The problem of our interest can be stated as follows. Suppose we have a generative model parameterized by $\theta$ that takes a set of design requirements $\vec{r}=\{r_1, r_2, ..., r_N\}$ as input and outputs a solution $x$ that addresses those requirements, e.g., $\pi_\theta(x | \vec{r})$. First, an engineer might want to prioritize a specific requirement $r_i$. Therefore, we aim to find a fine-tuned model $\pi_{\theta,r_i}$ such that a solution sampled from the model is optimal with respect to $r_i$. In some scenarios, there may be a new design requirement $n_j$ independent to the current generative model that an engineer would like the sampled solution to nevertheless satisfy. In such a case, the goal is to find a fine-tuned model $\pi_{\theta,n_j}$ from which a sampled solution would be optimal with respect to $n_j$.

Finally, given a set of prioritized requirements $\vec{p} \subset (\vec{r} \cup \vec{n})$, we aim to sample from the fine-tuned models 
a set of Pareto optimal solutions with respect to $\vec{p}$ that maximizes a Pareto-front quality, e.g., hypervolume \cite{zitzler1998multiobjective}.


\subsection{Illustrative example: GearFormer}

We use GearFormer, a recently developed generative model for gear train design \cite{etesam2024deep}, as an illustrative example for the current work. GearFormer is a Transformer-based model that takes multiple requirements as input via its encoder and outputs a gear train sequence via its decoder. The requirements it can handle are the speed ratio, output motion position, output motion direction, and input/output motion types. While it has been shown to outperform traditional search methods, an engineer does not have an option to express a preference of emphasizing one requirement over another, or explore multi-requirement trade-offs. Our goal is to fine-tune this model with respect to specific requirements and use the fine-tuned models to generate a high-quality Pareto front. 

We consider the two types of requirements as expressed in the problem definition. \emph{Original requirements} are those used to train GearFormer and therefore are used as an input to condition the output design. Note that these requirements are treated as equality constraints, i.e., they are target values such as speed ratio or output motion position that the design must meet. \emph{New requirements} are those never seen by the model during training. We consider metrics such as the bounding box volume and design cost, which can be evaluated given a design. In contrast to the original requirements, we intentionally chose new requirements formulated as inequality constraints, i.e., an engineer will be willing to accept any value that is below the specified bound value. 

\subsection{Challenges of fine-tuning a generative design model} 

A generative model for engineering design such as GearFormer is trained with a synthetic dataset of ($\vec{r}=$ requirements, $x=$ solution) pairs, where $x$ can be first generated using some rules and $\vec{r}$ is evaluated using a simulator for the generated $x$. The model is then trained to predict $x$ given $\vec{r}$, which means that the model has only seen designs that perfectly address the requirements. Therefore, during fine-tuning, any design that does not perfectly address a particular original requirement would likely deteriorate the performance of the pre-trained model. We show this effect in an ablation study presented in Experiments.

Now, suppose a RL-based method is used to fine-tune a pre-trained model with a typical policy gradient loss
\begin{equation}
\mathcal{L}(\theta) = -\mathbb{E}_{x \sim \pi_{\theta}} \left[ \log \pi_{\theta}(x|r_i) \mathcal{R}(x, r_i) \right]
\label{eq:policy gradient}
\end{equation}
where $\mathcal{R}$ is the reward for the solution sampled from the current policy. Since we aim to avoid degrading the policy with low-quality data, $\mathcal{R}$ can simply become a binary function that gives $1$ for an $x$ that satisfies $r_i$ and $0$ otherwise, i.e., equivalent to simply rejecting the sample. This means that with rejection sampled data, the \cref{eq:policy gradient} simply becomes a typical log probability loss used for supervised fine-tuning, e.g.,
\begin{equation}
\mathcal{L}(\theta) = -\mathbb{E}_{x' \sim \pi_{\theta}} \left[ \log \pi_{\theta}(x'|r_i) \right]
\label{eq:sft gradient}
\end{equation}
where $x'$ are solutions that satisfy the target value $r_i$. We therefore assume that SFT with rejection sampled data suffices as the fine-tuning step for original design requirements. 


However for new design requirements, the fine-tuning scenario is very similar to the one with LLMs. The pre-trained model is not trained on the new requirement data and does not have any sense of which output is good or bad with respect to the requirement. Therefore, a solution that does not perfectly satisfy the new requirement but is reasonably close can still provide useful signals for the model. We can therefore use a continuous reward value that reflects the degree of constraint violation. Based on these observations, a similar two-step technique used for fine-tuning LLMs such as using DPO or PPO with \emph{simulation feedback} in addition to SFT can be considered.

\subsection{Challenges of generating a good Pareto front}  

Randomly sampling a generative design model multiple times likely would not result in a good Pareto front because generation is not conditioned on different requirement preferences. One could sample multiple models each fine-tuned for different requirements, but you may get clusters of solutions at the extremes of only satisfying each requirement.

\section{Methods}

\begin{figure*}[t!]
    \centering
    \subfigure[SimFT method for original requirements -- SFT with simulation-augmented data]{\label{fig:a}\includegraphics[width=\textwidth]{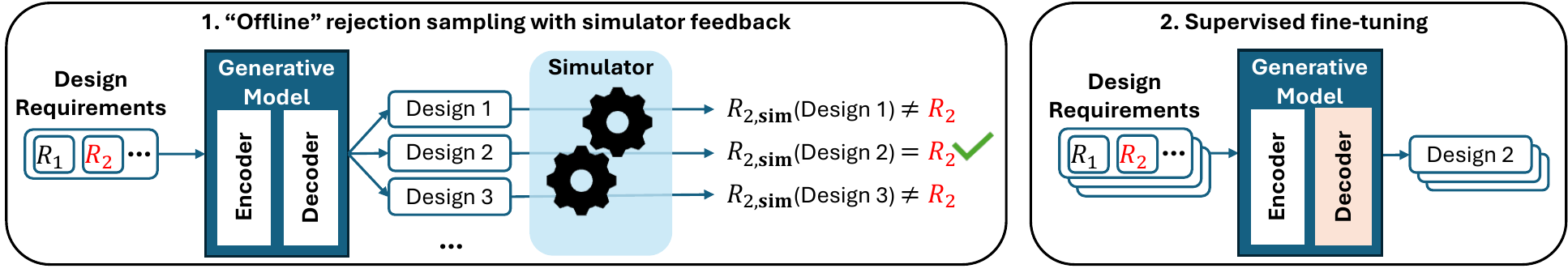}}
    
    \vskip\baselineskip

    \subfigure[SimFT method for new requirements -- Step 1: SFT with simulation augmented data]{\label{fig:b}\includegraphics[width=\textwidth]{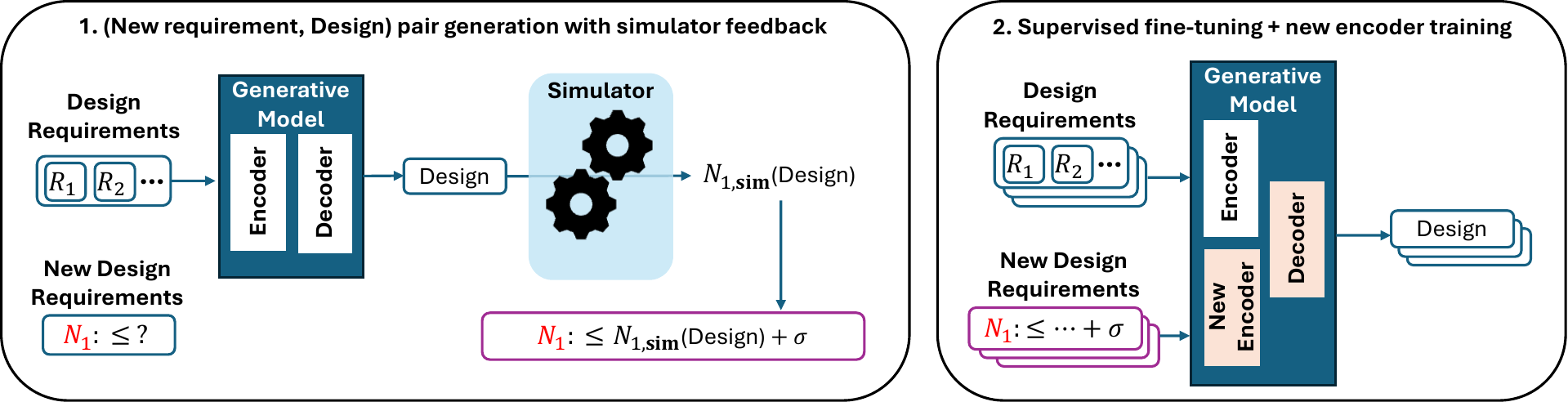}}

    \vskip\baselineskip

    \subfigure[SimFT method for new requirements -- Step 2, option 1: DPO with simulation augmented preference dataset]{\label{fig:c}\includegraphics[width=\textwidth]{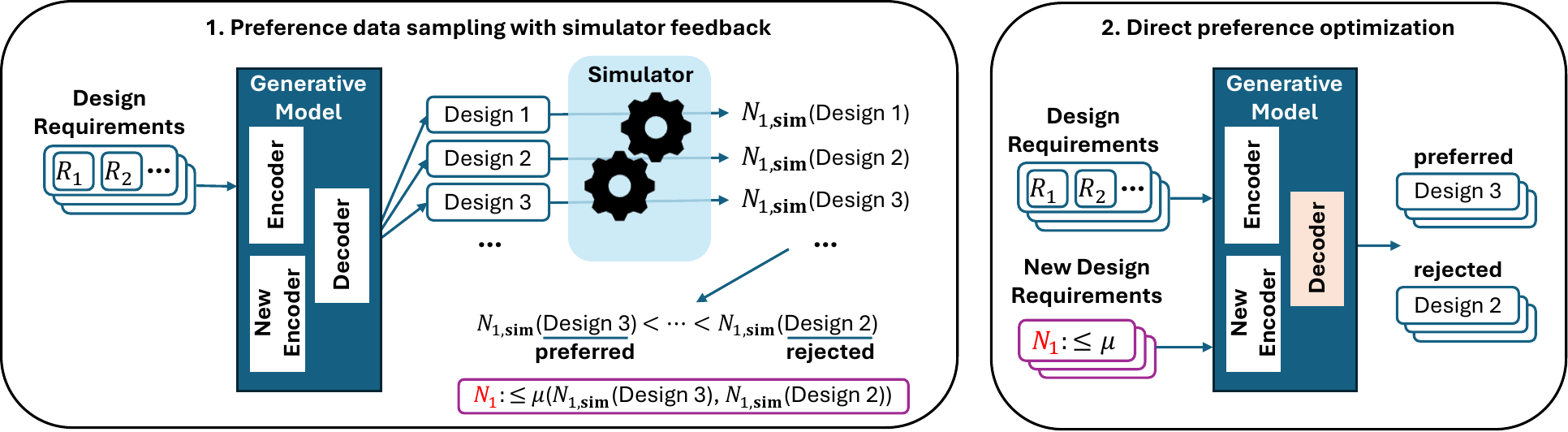}}

    \subfigure[SimFT method for new requirements -- Step 2, option 2 : PPO with a simulator to compute rewards.]{\label{fig:d}\includegraphics[width=0.85\textwidth]{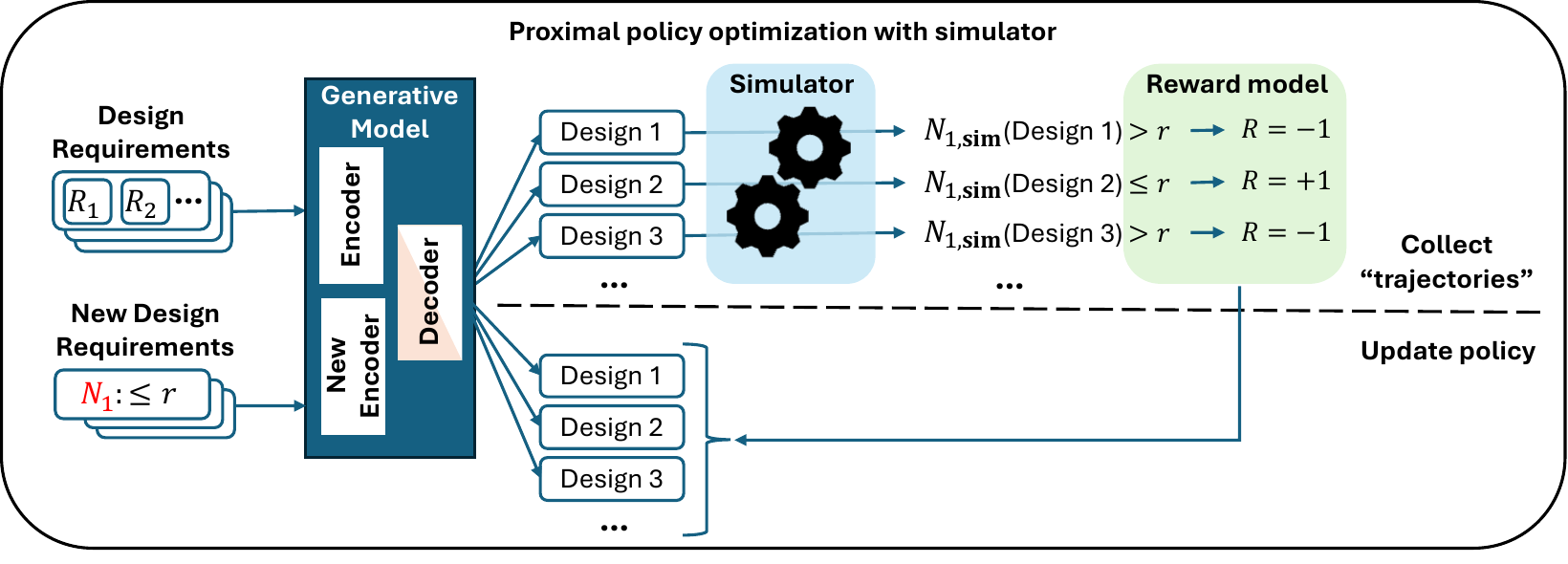}}
    
    \caption{SimFT data generation and training methods. }
    \label{fig:simft}
\end{figure*}

Given a pre-trained model that takes in a list of requirements and outputs a design that addresses those requirements, we first aim to fine-tune the model to prioritize satisfying a specific requirement. The fine-tuning methods are named \emph{SimFT}, where we use a physics simulator instead of human feedback to either generate the fine-tuning dataset or provide reward signals during PPO fine-tuning. See \cref{fig:simft} for illustration of all SimFT methods.

\subsection{SimFT methods for original requirements}

For an original requirement $r_i$, we perform a single SFT step as justified in the previous section. We generate an additional dataset for SFT by prompting the pre-trained model with a list of objectives (one of which is $r_i$), evaluate the designs generated using a simulator, and keep only the designs that meet $r_i$. This can be thought as synthetic data generation via rejection sampling, but performed in an offline mode before training. Also, we have the advantage of using a simulator to accurately evaluate the data generated and keeping only the perfect design solutions. 

The pre-trained model is fine-tuned with this dataset by minimizing the log probability loss (\cref{eq:sft gradient}). Note that we freeze the encoder while updating the decoder only.

\subsection{SimFT methods for new requirements}

For this category, a standard two-step fine-tuning methods developed for LLMs can be applied.

\paragraph{SFT.} In the first SFT stage, we generate the dataset in the following manner. We sample a design solution from the pre-trained model and compute the corresponding new requirement value using a simulator. We increase this value by some random variance and set it as the bound value for the requirement, synthetically generating a pair of the constraint bound value and a solution that satisfies the constraint.

We fine-tune the pre-trained model using this dataset by minimizing the log probability loss (\cref{eq:sft gradient}) but conditioned on $n_i$. We freeze the original encoder while training the decoder and also a new encoder that can take in the new requirement bound value.

Next, we apply either DPO or PPO to further fine-tune the model with respect to the new objectives. 

\paragraph{DPO.} The preference dataset is generated by sampling a pair of solutions from the pre-trained model and using a simulator to evaluate the requirement values. The solution with the lower requirement value is labeled as preferred while the other as rejected. We then assume the mean of the two requirement values as the constraint bound value, i.e., the preferred solution would satisfy the bound value while the rejected solution would not. We freeze both the original encoder and the new encoder while updating the decoder by minimizing the DPO loss \cite{rafailov2024direct}: 
\begin{flalign}
\mathcal{L}_{\text{DPO}}(\theta) = 
& -\mathbb{E}_{(x_w, x_l, n_i) \sim \mathcal{D}} 
\Bigg[
\log \sigma \Bigg( \beta
\log \frac{\pi_{\theta}(x_w \mid n_i)}{\pi_{\text{old}}(x_w \mid n_i)} & \nonumber \\
& - \beta \log \frac{\pi_{\theta}(x_l \mid n_i)}{\pi_{\text{old}}(x_l \mid n_i)} \Bigg) 
\Bigg]
\end{flalign}
where $x_w$ and $x_l$ are the preferred and rejected solutions for the requirement $n_i$. $\beta$ is the KL divergence penalty parameter and $\pi_{\text{old}}$ is the reference policy.

\paragraph{PPO with a simulator.} Another approach we can employ is PPO \cite{schulman2017proximal}, using the simulator to compute accurate rewards for each solution during exploration. For the loss, we use the clipped policy ratio with the KL divergence penalty:
\begin{flalign}
\mathcal{L}_{\text{RL}}(\theta) = 
& -\mathbb{E}_{(x, n_i) \sim \mathcal{D}} 
\Bigg[
\min \Big( \frac{\pi_{\theta}(x \mid n_i)} {\pi_{\text{old}}(x \mid n_i)} \mathcal{R}(x, n_i), & \nonumber \\
& \text{clip}(\frac{\pi_{\theta}(x \mid n_i)} {\pi_{\text{old}}(x \mid n_i)}, 1 - \epsilon, 1 + \epsilon) 
\mathcal{R}(x, n_i)
\Big) & \nonumber \\ &  - \beta \text{KL}\big(\pi_{\theta}, \pi_{\text{old}} \big)
\Bigg]
\end{flalign}
where $\mathcal{R}(x, n_i)$ is a reward function computed using the simulator output and normalized to $[-1, 1]$, i.e., $\mathcal{R}=1$ if $x$ is evaluated to meet the bound value $n_i$ and approaches $-1$ as the violation increases (See \cref{AppA} for details).

\subsection{epsilon-sampling for Pareto-front generation}

\begin{figure}[h!]
    \centering
    \includegraphics[width=0.98\linewidth]{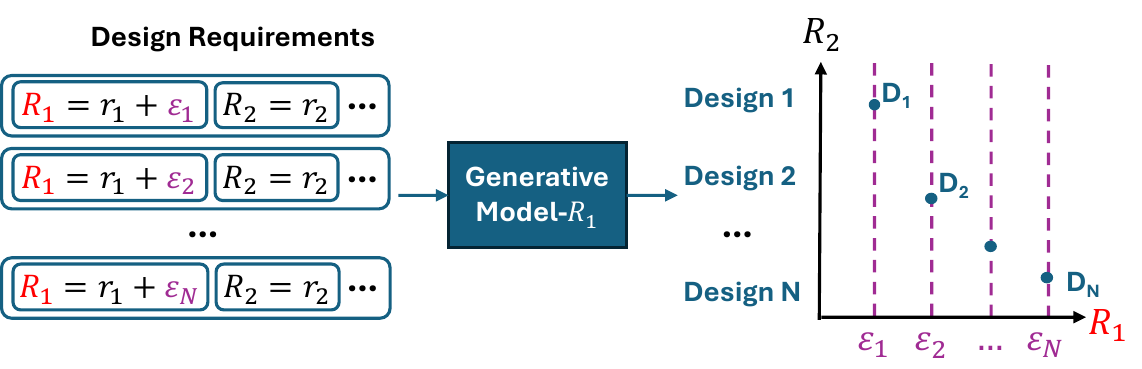}
    \caption{Epsilon-sampling with a SimFT model. The target requirement values for $R_1$ is incremented with $\epsilon_i$ and a SimFT model for $R_1$ is sampled to construct a Pareto front.}
    \label{fig:epsilon}
\end{figure}

We propose epsilon-sampling (\cref{fig:epsilon}) inspired by the epsilon-constraint method \cite{haimes1971bicriterion} to obtain Pareto-optimal solutions with SimFT models. The epsilon-constraint method is a well-known technique to produce a Pareto front with gradient-based algorithms for multi-objective optimization problems. Given a pair of objectives, the method sets one objective as a constraint and solves multiple single-objective constrained optimization problems by incrementing the threshold value $\epsilon$ imposed on the constraint. Solutions to these problems form a Pareto front.

We apply this idea for sampling generative models to construct a Pareto front. Given a set of requirements $\vec{r}$, we assume that a model fine-tuned for $r_i$ can best enforce that constraint; therefore, sampling from that model would be equivalent to posing $r_i$ as a constraint and the rest of requirements as objectives. We sample multiple solutions from this model by varying the target value by $r_i \pm \epsilon$, effectively mimicking the epsilon-constraint method. The same technique can be applied with new requirements $n_i$. 

\section{Experiments}

We evaluate the performance improvements made by SimFT methods and perform ablation studies to elucidate important aspects of SimFT. We then evaluate e-SimFT against several baselines in generating high-quality Pareto fronts.

\subsection{Experimental setups}

\paragraph{Pre-trained model and simulator.}

We use GearFormer \cite{etesam2024deep} as the pre-trained model. We also use the simulator developed for GearFormer and extend it to compute the new requirements required for this work. While it would be ideal to test our method on multiple generative models, GearFormer was the only work that provided the model, simulator, and dataset required for our experiments.

\paragraph{Design requirements.}

We consider four design requirements. Two original requirements of GearFormer, \textbf{speed} ratio and output motion \textbf{position}, are posed as equality constraints. The other two are new requirements that were not considered in training GearFormer, bounding box volume (\textbf{b.box}) and design \textbf{cost}, posed as inequality constraints. 

\paragraph{Design scenarios.}

We generated new 30 random test problems based on the distributions of requirement metrics obtained from the original GearFormer dataset \cite{etesam2024deep}. For each test problem, we consider 10 different trade-off scenarios -- the all possible two-way and three-way combinations of the four requirements under consideration. For each design scenario, a sample budget of $\mathcal{N}$ is assumed. How this budget is used varies depending on the methods employed, as explained in the following section.

\paragraph{Baselines.}

The first baseline involves sampling the pre-trained GearFormer $\mathcal{N}$ times for each design scenario.

Two distinct and recent multi-objective alignment methods are chosen as additional baselines. First, \emph{Rewarded Soup} (RS) \cite{rame2024rewarded} linearly interpolates the weights of models fine-tuned for each specific objective, given the preference weights assigned for each objective. A Pareto front can be constructed by sampling from multiple of these linearly interpolated models with varying preference weights. We define $\mathcal{N}_s$ combinations of preference weights and allocate $\mathcal{N}/\mathcal{N}_s$ sampling budget for each combination.

We also chose \emph{Rewards-in-Context} (RiC) \cite{yang2024rewards}. RiC performs SFT on the pre-trained model with outputs associated with their reward/preference values, which are encoded as additional input to the model. For this work, we train a new encoder that can take in preference values for each requirement, indicating which requirements to prioritize. We define $\mathcal{N}_r$ combinations of requirement preferences and allocate $\mathcal{N}/\mathcal{N}_r$ sampling budget for each combination. Baseline implementation details including the preference weight combinations used can be found in \cref{AppA}.

\paragraph{e-SimFT.} Relevant SimFT models are chosen based on the design scenario and we allocate $\mathcal{N}/2$ or $\mathcal{N}/3$ (for two- or three-requirements) sampling budget to each model. We then create an evenly spaced values of $\epsilon$, sized either $\mathcal{N}/2$ or $\mathcal{N}/3$, within [-5, 5] for original requirements and [0, 10] for new requirements, and add these values to the requirement values before sampling SimFT models.

We also test two conditions as ablation: SimFT only or epsilon-sampling only. For the former, we use the same sampling budget allocation as e-SimFT with relevant SimFT models but do not employ epsilon-sampling. For the latter, we follow the same epsilon-varying schedule as the e-SimFT method but use the pre-trained GearFormer model. 

\paragraph{Evaluation metric.}

For each requirement, the degree of constraint violation for the design solutions generated by different models is normalized as $[0,1]$. These values are used to determine Pareto optimal solutions for each problem and the hypervolume of the Pareto front \cite{zitzler1998multiobjective} is used to compare e-SimFT versus other baselines. 

\paragraph{Dataset and training}

We use the validation and test portion of the original GearFormer dataset \cite{etesam2024deep} for all our fine-tuning and testing, $|D|=7360$, which is around 1\% of the training dataset used for GearFormer. We believe that this is a reasonable ratio of training versus fine-tuning data for generative design models in practice. Details on the dataset and training can be found in the \cref{AppA}.

\addtocounter{figure}{1} 
\begin{figure*}[b!]
    \centering
    \begin{subfigure}
        \centering
        \includegraphics[width=0.9\textwidth]{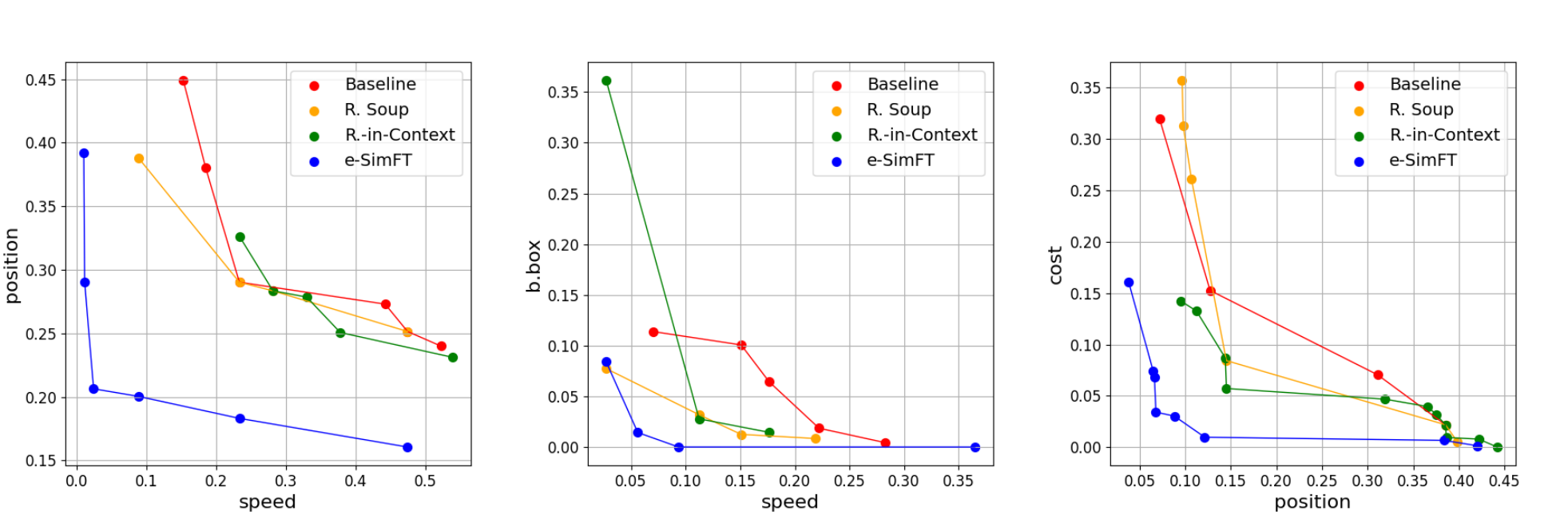}
    \end{subfigure}
        
    \begin{subfigure}
        \centering
        \includegraphics[width=0.9\textwidth]{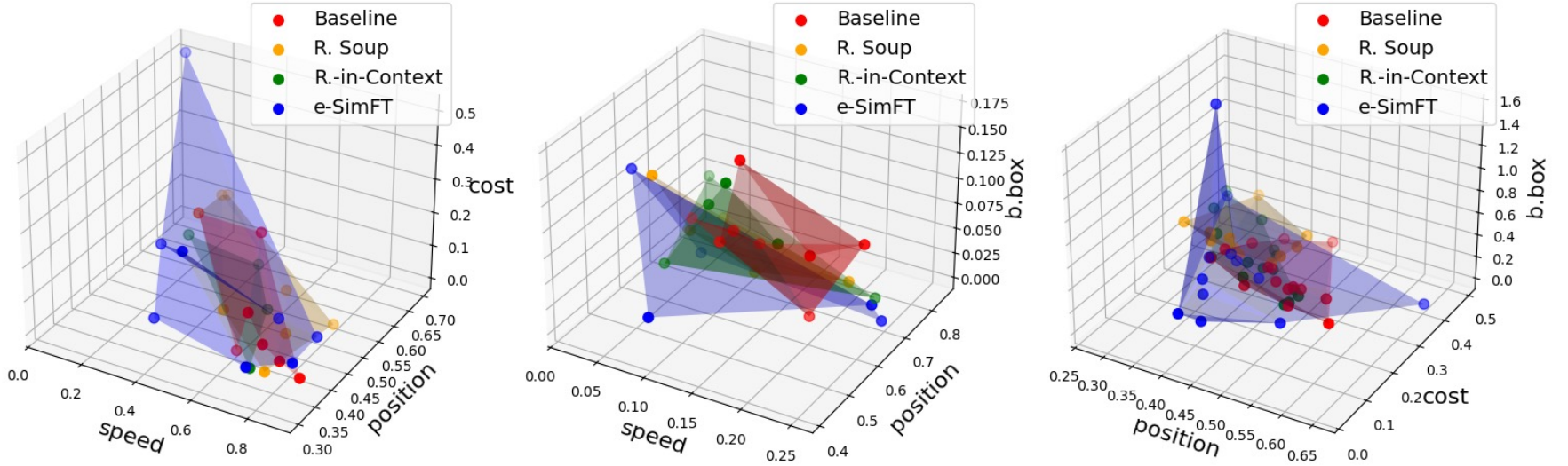}
    \end{subfigure}
    
    \caption{Pareto fronts generated by e-SimFT versus other methods for sample design problems.}
    \label{fig:pareto_fronts}
\end{figure*}
\addtocounter{figure}{-2} 

\subsection{Evaluation of SimFT methods}

\begin{figure}[h!]
    \centering
    \includegraphics[width=1\linewidth]{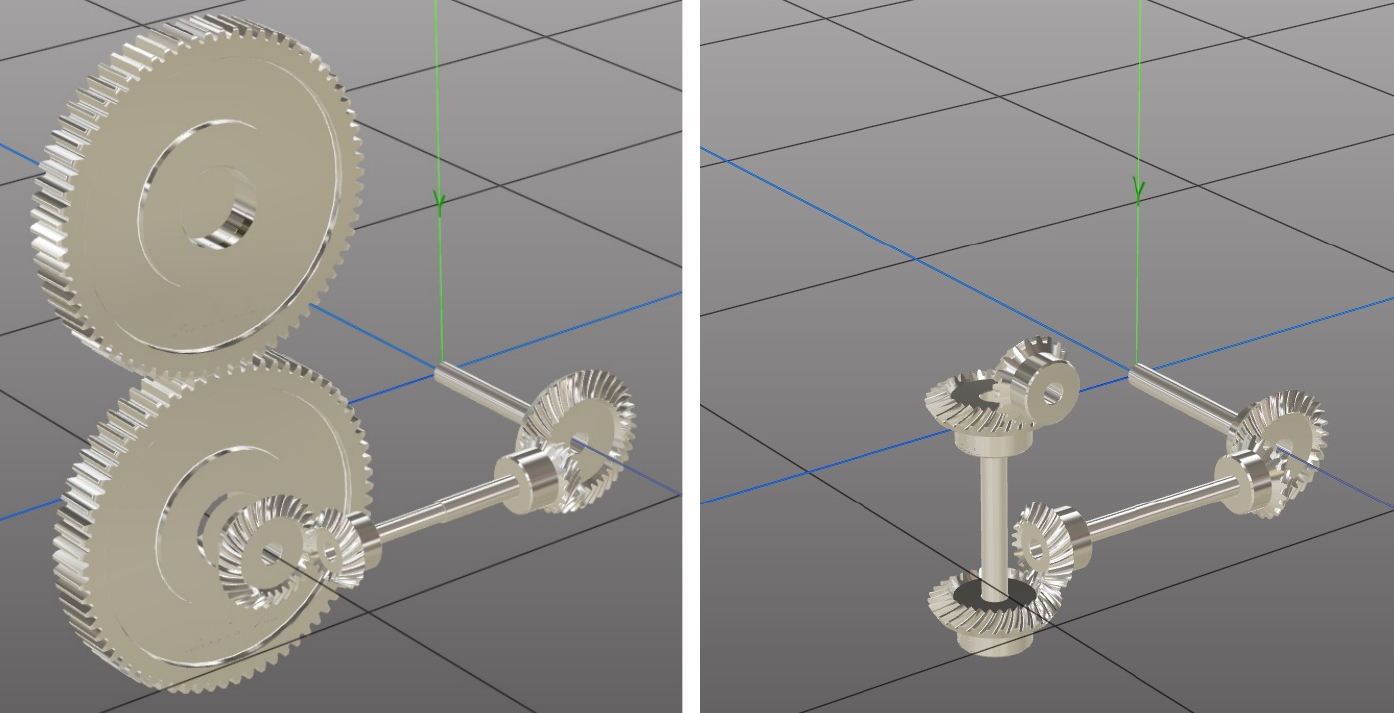}
    \caption{Example of gear designs produced for a sample problem with the original GearFormer versus a SimFT model fine-tuned for bounding box volume. The first design has a volume of 0.018$m^3$ while the second design has a much lower volume of 0.008$m^3$.}
    \label{fig:bb}
\end{figure}

\cref{tab:simft} presents the improvements made by SimFT methods for each requirement. For original requirements (speed and position), the metrics are based on \citeauthor{etesam2024deep}. For new requirements (cost and bounding box volume), we calculate the percentage of the problems from the test dataset for which the first solution generated by the original GearFormer or SimFT models satisfy the respective requirement. We observed that DPO performed slightly better than RL for both requirements. As an example, \cref{fig:bb} shows solutions obtained for the same design problem with the original GearFormer model vs. a SimFT model.
\begin{table}[h!]
\small
\centering
\caption{Performance improvements by SimFT methods}
\begin{tabular}{
  lccc
}
\toprule
 \textbf{Requirment} & \textbf{Baseline} & \multicolumn{2}{c}{\textbf{SimFT method}} \\
 &  & SFT & DPO / RL via PPO \\
\midrule
speed [log(·)] $\downarrow$ & $0.0171$ & $0.0139$ & N/A \\
position [m] $\downarrow$ & $0.0338$ & $0.0317$ & N/A \\
cost $\uparrow$ & $52.8\%$ & $54.1\%$ & $66.9\%$ / $65.4\%$\\
b.box $\uparrow$ & $49.4\%$ & $55.2\%$ & $62.3\%$ / $59.1\%$ \\
\bottomrule
\end{tabular}
\label{tab:simft}
\end{table}

\paragraph{Importance of offline rejection sampling for SFT.}

To show the importance of using only the samples that satisfy the equality constraint requirements, we ran an ablation study where any new sample drawn was accepted for the SFT training data. The performances for the speed and position requirements dropped to 0.0179 and 0.0339, compared to the baseline performances of 0.0171 and 0.0338. 

\paragraph{Performance trade-offs during DPO.}

\begin{figure}[h!]
    \centering
    \includegraphics[width=0.95\linewidth]{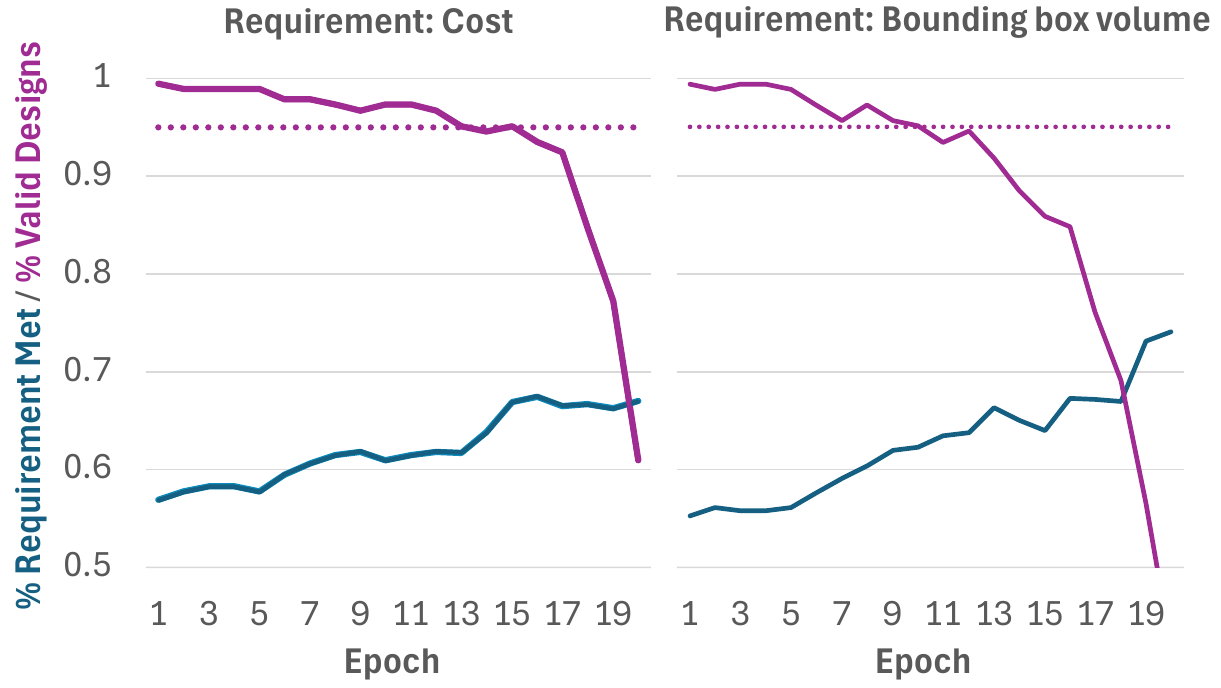}
    \caption{DPO improves the percentage of requirements met at the expense of the percentage of valid designs.}
    \label{fig:perf}
\end{figure}

Because DPO is performed using a preference loss that differs from the original cross-entropy loss used for the pre-trained model, over-training the model can lead to significant deterioration of its original performance. \cref{fig:perf} shows that after 16 and 12 epochs, respectively, the percentage of valid designs produced by GearFormer drops below 95\% (the performance reported in \cite{etesam2024deep}) and at a significant rate in the subsequent epochs. Note that the DPO training loss and the requirement-met performance continues to improve over these epochs. Therefore, the best model checkpoint should be picked after confirming that other performances of the model have not deteriorated below the required thresholds.
\begin{table*}[b!]
\caption{Comparison of Pareto-front hypervolumes. Number of test problems = 30 and samples = 30 per scenario.}
\small
\centering
\begin{tabular}{
  lccccccc
}
\toprule
 \textbf{Design Scenario}& \multicolumn{6}{c}{\textbf{Method}} \\
 & Baseline & R. Soup & R.-in-Context & SimFT & e-sample & e-SimFT \\
\midrule
speed : position & $0.564^{\pm0.188}$ & $0.588^{\pm0.196}$ & $0.573^{\pm0.211}$ & $0.591^{\pm0.201}$ & $0.683^{\pm0.151}$ & $\mathbf{0.699}^{\pm0.162}$ \\
speed : cost & $0.611^{\pm0.233}$ & $0.610^{\pm0.227}$ & $0.603^{\pm0.234}$ & $0.612^{\pm0.244}$ & $0.657^{\pm0.240}$ & $\mathbf{0.666}^{\pm0.228}$ \\
speed : b.box & $0.712^{\pm0.192}$ & $0.715^{\pm0.274}$ & $0.715^{\pm0.202}$ & $0.749^{\pm0.226}$ & $0.739^{\pm0.225}$ & $\mathbf{0.810}^{\pm0.198}$ \\
position : cost & $0.458^{\pm0.224}$ & $0.479^{\pm0.227}$ & $0.483^{\pm0.246}$ & $0.474^{\pm0.225}$ & $0.470^{\pm0.240}$ & $\mathbf{0.492}^{\pm0.253}$ \\
position : b.box & $0.524^{\pm0.215}$ & $0.455^{\pm0.216}$ & $0.527^{\pm0.218}$ & $0.507^{\pm0.196}$ & $\mathbf{0.555}^{\pm0.201}$ & $0.537^{\pm0.190}$ \\
cost : b.box & $0.571^{\pm0.256}$ & $0.510^{\pm0.277}$ & $\mathbf{0.580}^{\pm0.245}$ & $0.555^{\pm0.248}$ & $0.516^{\pm0.273}$ & $0.521^{\pm0.283}$ \\
\cmidrule{2-7}
\textbf{mean} & $0.573^{\pm0.219}$ & $0.560^{\pm0.238}$ & $0.580^{\pm0.227}$ & $0.582^{\pm0.224}$ & $0.603^{\pm0.225}$ & $\mathbf{0.621}^{\pm0.223}$ \\
\midrule
speed : position: cost & $0.375^{\pm0.217}$ & $0.395^{\pm0.232}$ & $0.402^{\pm0.236}$ & $0.427^{\pm0.228}$ & $0.478^{\pm0.242}$ & $\mathbf{0.480}^{\pm0.217}$ \\
speed : position : b.box & $0.435^{\pm0.212}$ & $0.435^{\pm0.212}$ & $0.431^{\pm0.198}$ & $0.505^{\pm0.189}$ & $0.502^{\pm0.179}$ & $\mathbf{0.526}^{\pm0.200}$ \\
speed : cost : b.box & $0.466^{\pm0.244}$ & $0.471^{\pm0.260}$ & $0.500^{\pm0.234}$ & $\mathbf{0.542}^{\pm0.229}$ & $0.480^{\pm0.237}$ & $0.520^{\pm0.246}$ \\
position : cost : b.box & $0.337^{\pm0.203}$ & $0.312^{\pm0.206}$ & $0.375^{\pm0.226}$ & $\mathbf{0.380}^{\pm0.203}$ & $0.354^{\pm0.235}$ & $0.366^{\pm0.212}$ \\
\cmidrule{2-7}
\textbf{mean} & $0.403^{\pm0.219}$ & $0.403^{\pm0.229}$ & $0.427^{\pm0.224}$ & $0.463^{\pm0.213}$ & $0.454^{\pm0.225}$ & $\mathbf{0.473}^{\pm0.219}$ \\
\bottomrule
\end{tabular}
\label{tab:hv}
\end{table*}

\begin{table*}[b!]
\caption{Comparison of Pareto-front hypervolumes. Number of test problems = 30 and samples = 300 per scenario.}
\small
\centering
\begin{tabular}{
  lccccccc
}
\toprule
 \textbf{Design Scenario}& \multicolumn{6}{c}{\textbf{Method}} \\
 & Baseline & R. Soup & R.-in-Context & SimFT & e-sample & e-SimFT \\
\midrule
speed : position & $0.680^{\pm0.169}$ & $0.696^{\pm0.163}$ & $0.680^{\pm0.186}$ & $0.702^{\pm0.164}$ & $0.769^{\pm0.128}$ & $\mathbf{0.775}^{\pm0.137}$ \\
speed : cost & $0.712^{\pm0.220}$ & $0.716^{\pm0.218}$ & $0.719^{\pm0.213}$ & $0.718^{\pm0.223}$ & $0.762^{\pm0.190}$ & $\mathbf{0.776}^{\pm0.198}$ \\
speed : b.box & $0.828^{\pm0.147}$ & $0.836^{\pm0.200}$ & $0.866^{\pm0.138}$ & $0.874^{\pm0.151}$ & $0.866^{\pm0.150}$ & $\mathbf{0.912}^{\pm0.131}$ \\
position : price & $0.584^{\pm0.211}$ & $0.575^{\pm0.224}$ & $0.574^{\pm0.234}$ & $0.588^{\pm0.227}$ & $0.589^{\pm0.243}$ & $\mathbf{0.592}^{\pm0.243}$ \\
position : b.box & $0.625^{\pm0.202}$ & $0.558^{\pm0.207}$ & $0.633^{\pm0.199}$ & $0.621^{\pm0.193}$ & $\mathbf{0.676}^{\pm0.188}$ & $0.651^{\pm0.188}$ \\
cost : b.box & $0.723^{\pm0.229}$ & $0.622^{\pm0.264}$ & $\mathbf{0.703}^{\pm0.237}$ & $0.689^{\pm0.250}$ & $0.676^{\pm0.257}$ & $0.666^{\pm0.274}$ \\
\cmidrule{2-7}
\textbf{mean} & $0.692^{\pm0.199}$ & $0.667^{\pm0.215}$ & $0.696^{\pm0.204}$ & $0.698^{\pm0.205}$ & $0.723^{\pm0.198}$ & $\mathbf{0.729}^{\pm0.202}$ \\
\midrule
speed : position : cost & $0.508^{\pm0.227}$ & $0.501^{\pm0.236}$ & $0.516^{\pm0.239}$ & $0.527^{\pm0.234}$ & $0.574^{\pm0.219}$ & $\mathbf{0.589}^{\pm0.226}$ \\
speed : position: b.box & $0.568^{\pm0.190}$ & $0.521^{\pm0.211}$ & $0.579^{\pm0.206}$ & $0.591^{\pm0.192}$ & $0.652^{\pm0.173}$ & $\mathbf{0.663}^{\pm0.184}$ \\
speed : cost : b.box & $0.600^{\pm0.230}$ & $0.597^{\pm0.251}$ & $0.620^{\pm0.227}$ & $0.649^{\pm0.234}$ & $0.655^{\pm0.234}$ & $\mathbf{0.687}^{\pm0.233}$ \\
position : cost : b.box & $0.472^{\pm0.218}$ & $0.406^{\pm0.229}$ & $0.491^{\pm0.232}$ & $0.497^{\pm0.230}$ & $\mathbf{0.518}^{\pm0.229}$ & $0.500^{\pm0.247}$ \\
\cmidrule{2-7}
\textbf{mean} & $0.537^{\pm0.217}$ & $0.506^{\pm0.232}$ & $0.551^{\pm0.226}$ & $0.566^{\pm0.223}$ & $0.600^{\pm0.215}$ & $\mathbf{0.610}^{\pm0.224}$ \\
\bottomrule
\end{tabular}
\label{tab:hv_300}
\end{table*}

\paragraph{Using binary reward function for PPO.}

Considering the new requirements are inequality constraints, we could implement a binary reward function for PPO that simply assigns a score of 1 or -1 if the requirement is met or not. We found that fine-tuning the model with this reward function for 20 epochs, the best accuracies were 62.3\% for the cost requirement and 60.4\% for the bounding box volume requirement, versus 65.4\% and 59.1\% reported in \cref{tab:simft}.

\subsection{Evaluation of e-SimFT for Pareto-front generation}

Finally, we report the hypervolumes of Pareto fronts obtained with e-SimFT and other methods in Tables \ref{tab:hv} and \ref{tab:hv_300}. We also present Pareto fronts generated by different methods for sample design problems in \cref{fig:pareto_fronts}.

We considered two settings for the hypervolume comparison. In the first setting, we used the sampling budget of $\mathcal{N}=30$, which for each design scenario takes about 90 seconds of inference and simulation time on a machine with a Tesla V100-SXM2-16GB GPU and a AMD EPYC 16-core processor. This is assumed to be a relatively short time that an engineer needs to wait to obtain a Pareto front with multiple optimal solutions. In the second setting, we set $\mathcal{N}=300$ to examine how each method scales with increased sampling.

\cref{tab:hv} shows that on average, e-SimFT is the best method for both two- and three-requirements scenarios. One could also observe that on average, SimFT and epsilon-sampling on their own do not perform as well as e-SimFT. Only in one scenario an alternative multi-objective alignment method, RiC, achieved the highest hypervolume. 

\cref{tab:hv_300} shows a similar pattern when the sampling size is increased by 10 times. On average, e-SimFT is still found to be the best for both two- and three-requirements scenarios.

\section{Conclusions}

This work introduces e-SimFT, a new framework for Pareto-front design exploration with simulation fine-tuned generative models in engineering design. It employs multiple preference alignment methods, named SimFT methods, by using a simulator to fine-tune a generative model with respect to a specific requirement prioritized by an engineer. SimFT models are then sampled using the epsilon-sampling method to construct a high-quality Pareto front for design scenarios involving trade-offs among multiple requirements. In both two- and three-requirements scenarios, e-SimFT outperformed latest multi-objective alignment methods in terms of the hypervolumes of the Pareto fronts generated.

We believe that many parallels exist between generative AI and engineering design as both domains strive for creative automation. This work showcases an innovative application of generative AI research to facilitate engineering design exploration.

\clearpage
\newpage



\bibliography{ref}
\bibliographystyle{icml2025}

\clearpage
\newpage
\appendix
\onecolumn

\section{Experiment Details}
\label{AppA}

\subsection{Dataset}

We use the validation and test portion of the original GearFormer dataset \cite{etesam2024deep} for all our fine-tuning and testing, $|D|=7360$. 5\% of the dataset was withheld for testing SimFT methods including their ablation studies. The rest was used for fine-tuning, i.e., $D_{\text{ft}} \cup D_{\text{test}} = D, |D_{\text{ft}}|=6992, |D_{\text{test}}|=368$.

For the original requirement SimFT, use the whole $D_{\text{ft}}$ for SFT; where 90\% is used for training and 10\% is held for validation. 

For the new requirement SimFT, use the first half $D_{\text{ft},1}$ for SFT and the other half $D_{\text{ft},2}$ for DPO/RL, i.e., $D_{\text{ft},1} \cup D_{\text{ft},2} = D_{\text{ft}}, |D_{\text{ft},i}|=3496$. Again for each subset, we use 90\% for training and 10\% for validation.

\subsection{Training SFT models.}

\paragraph{SFT for original requirements:} The training was performed until the validation loss increased. For the speed SimFT, it was stopped at epoch \#16, and for the position SimFT, it was stopped at epoch \#19. We used the learning rate of 1e-6 and the batch size of 64 for both.

\paragraph{SFT for new requirements:} The training was performed until the validation loss increased. For both the speed and position SimFT models, it was stopped at epoch \#6. We used the learning rate of 1e-5 (including the new encoder) and the batch size of 64 for both.

\paragraph{DPO for new requirements:} The training was done for 20 epochs and the model checkpoint with the best requirement improvement was picked \emph{post hoc}, subject to the criterion that 95\% of the generated solutions are valid. For the cost SimFT, this was at epoch \#15 while for the boundinb box volume SimFT, it was at epoch \#10. We used the learning rate of 1e-6, $\beta=0.1$, and the batch size of 64 for both.

\paragraph{PPO for new requirements:} Same as DPO, the training was done for 20 epochs and the model checkpoint with the best requirement improvement was picked \emph{post hoc}, subject to the criterion that 95\% of the generated solutions are valid. For the cost SimFT, this was at epoch \#9 while for the boundinb box volume SimFT, it was at epoch \#4. We used the learning rate of 1e-5, $\beta=0.1$, and the batch size of 64 for both.

For the reward function, we used the following normalization function based on the evaluation of each solution using a simulator:
\[
    \mathcal{R}(x)= 
\begin{cases}
    1,& \text{if } \Tilde{n}(x)\leq n\\
    1-\frac{2(\Tilde{n}(x)-n)}{1+\Tilde{n}(x)-n},              & \text{otherwise}
\end{cases}
\]
where $x$ is the solution, $n$ is the target requirement value, and $\Tilde{n}(x)$ is the evaluated requirement value. Note that because we compute the exact reward for each solution as a whole, no actor-critic models are employed.

\subsection{Baseline implementations}

\paragraph{Rewarded Soup:}

For a given weight preference combination, we linearly interpolate the parameters (weights) of SimFT models to create a new model specific for that combination. For Pareto-front generation, we use the following weight combinations. For the two-requirement scenarios, $w_1=[0, 0.2, 0.4, 0.6, 0.8, 1]$ and $w_2 = 1 - w_1$. For the three-requirement scenarios, $w_1=[0, 0, 0, 0.33, 0.5, 0.5, 1]$, $w_2=[0, 0.5, 1, 0.33, 0, 0.5, 0]$, and $w_3= 1 - w_1 - w_2$.

\paragraph{Rewards-in-Context:} 

Using the problems defined in $D_{\text{ft}}$, the training data is generated by sampling a solution for each problem using GearFormer and determining whether the solution meets each of the four requirements of interest or not. Based on this, we can create a preference weight vector that indicates whether a particular requirement is met or not. We then perform supervised fine-tuning with this dataset using log probability loss, while training two new encoders -- one for encoding the new requirements and another for encoding the preference weight vector. We train until the validation loss increases, which was at epoch \#7 . We used the learning rate of 1e-6 for all models and the batch size of 64. For Pareto-front generation, we use the following weight combinations. For the two-requirement scenarios, $w_1=[0, 1, 1]$ and $w_2 = [1, 0, 1]$. For the three-requirement scenarios, 
$w_1=[0, 0, 1, 0, 1, 1, 1]$, $w_2=[0, 1, 0, 1, 0, 1, 1]$, and $w_3=[1, 0, 0, 1, 1, 0, 1]$.

\end{document}